\documentclass{article}

    \usepackage[preprint]{neurips_2025}

\usepackage[utf8]{inputenc} 
\usepackage[T1]{fontenc}    
\usepackage{hyperref}       
\usepackage{url}            
\usepackage{booktabs}       
\usepackage{amsfonts}       
\usepackage{nicefrac}       
\usepackage{microtype}      
\usepackage{xcolor}         

\usepackage{graphicx}
\usepackage{pifont}
\usepackage{multirow}
\usepackage{amsmath}
\usepackage{wrapfig}
\usepackage{caption}  
\usepackage{lipsum} 
\usepackage{floatflt}

\title{LAS: Loss-less ANN-SNN Conversion for Fully Spike-Driven Large Language Models}

\author{%
  Long Chen, Xiaotian Song, Yanan Sun \\
  Collage of Computer Science, Sichuan University\\
}

\begin{document}

\maketitle

\begin{abstract}
Spiking Large Language Models (LLMs) have emerged as an energy-efficient alternative to conventional LLMs through their event-driven computation. To effectively obtain spiking LLMs, researchers develop different ANN-to-SNN conversion methods by leveraging pre-trained ANN parameters while inheriting the energy efficiency of SNN. However, existing conversion methods struggle with extreme activation outliers and incompatible nonlinear operations of ANN-based LLMs. To address this, we propose a loss-less ANN-SNN conversion for fully spike-driven LLMs, termed LAS. Specifically, LAS introduces two novel neurons to convert the activation outlier and nonlinear operation of ANN-based LLMs. Moreover, LAS tailors the spike-equivalent Transformer components for spiking LLMs, which can ensure full spiking conversion without any loss of performance. Experimental results on six language models and two vision-language models demonstrate that LAS achieves loss-less conversion. Notably, on OPT-66B, LAS even improves the accuracy of 2\% on the WSC task. In addition, the parameter and ablation studies further verify the effectiveness of LAS.\footnote{Available code:~https://github.com/lc783/LAS}
\end{abstract}

\section{Introduction}

Large Language Models (LLMs), in recent years, have revolutionized artificial intelligence by achieving state-of-the-art performance in language processing~\cite{liu2024deepseek} and multimodal tasks~\cite{wang2024qwen2}. However, there exist significant challenges in the training and inference process of LLMs, particularly the computational complexity and unsustainable energy consumption. This gap has driven an urgent search for more efficient computing paradigms that can support the ever-growing scale of LLMs. Inspired by low-power biological neural systems, Spiking Neural Networks (SNNs) offer a promising alternative~\cite{bohte2000spikeprop,gerstner2014neuronal}. More specifically, SNNs use discrete, sparse spikes to encode and process information, which can significantly reduce energy consumption than traditional Artificial Neural Networks (ANNs)~\cite{davies2018loihi,duan2024memristor,yao2024spikeChip}.

To effectively build the SNN models, existing methods can be divided into direct training and ANN-to-SNN conversion. The former one uses surrogate gradients to overcome the challenges posed by the non-differentiable nature of spike events~\cite{8891809,zenke2021remarkable,song2024one}. However, the direct training method naturally suffers from huge computational costs, which are unaffordable for most researchers using this method to build large SNN models in practice. The later one, i.e., ANN-to-SNN conversion, involves converting pre-trained ANNs into SNNs by transferring their learned parameters into a spiking framework, thus preserving accuracy while benefiting from the energy efficiency of spike-based computation~\cite{cao2015spiking, rueckauer2016theory,rueckauer2017conversion}. Through ANN-to-SNN conversion, we can easily obtain the high-performance SNN models.

Although many conversion methods have been successfully applied to Convolutional Neural Networks (CNNs)~\cite{Deng2021OptimalCO,Li2022EfficientAA}, extending them to Transformer-based LLMs remains two main challenges. 
 \begin{wrapfigure}{r}{0.5\columnwidth}
  \centering
    \includegraphics[width=\linewidth]{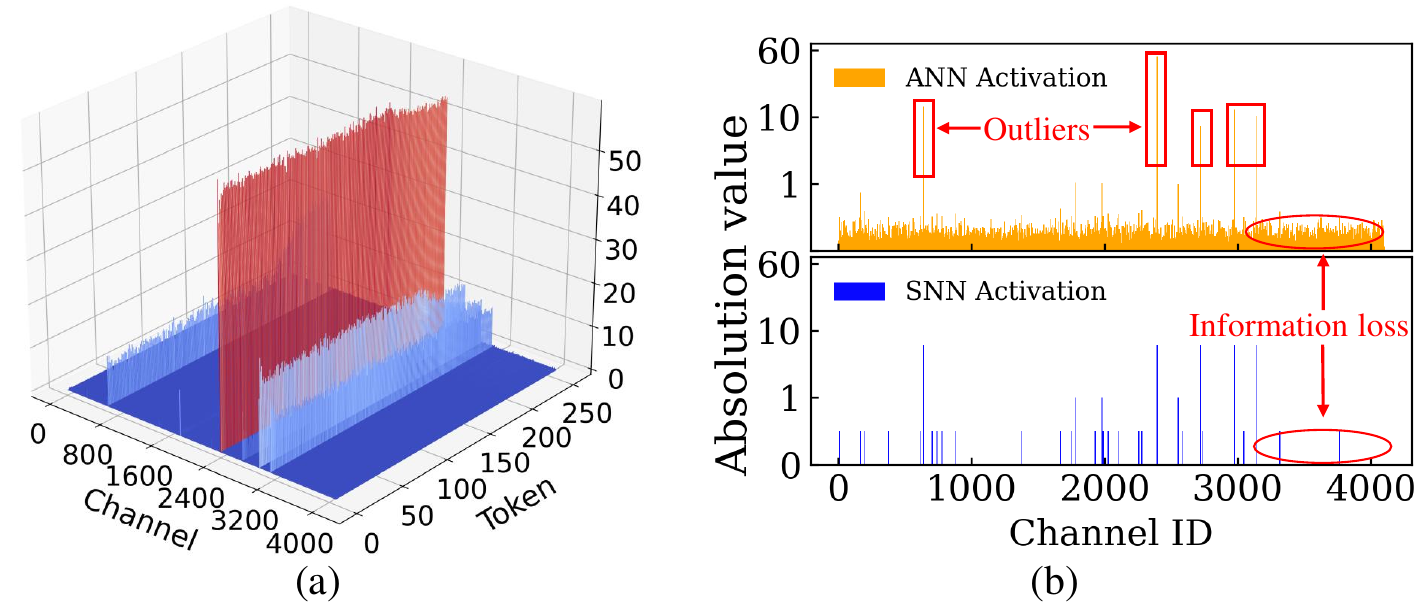}
    \caption{Visualizations of outliers on OPT-7B. (a) Extensive outliers from attention mechanism. (b) The information loss of the converted activations.}
  \label{hidenstate}
\end{wrapfigure}
First, as shown in Figure~\ref{hidenstate}, LLMs often exhibit activation outliers that significantly affect model performance. When these values are represented by spiking neurons, many activations are compressed into a narrow range, leading to severe information loss. Second, Transformer-based LLMs always have more complex architecture than CNNs. Specifically, LLMs depend on nonlinear operations, e.g., Self-Attention, LayerNorm, GELU, and Softmax. 
Unfortunately, accurately representing these components using the linear behavior of spiking neurons still remains a significant challenge.

To address these issues, we proposes a loss-less ANN-SNN Conversion for fully spike-driven LLMs, termed LAS. More specifically, to address activation outliers, we propose the Outlier-Aware Threshold neuron, which employs dual Multi-Threshold sub-neurons to process normal and outlier activations separately. Next, to approximate nonlinear operations, we introduce the Hierarchically Gated neuron, leveraging a hierarchical decomposition approximation through grouped spiking sub-neurons. Finally, we design the Spike-Equivalent LLM architecture, converting all key modules into spike-equivalent counterparts without converse error. Our contributions are summarized as follows:
\begin{itemize}
    \item \textbf{Two Novel Neurons.} We propose the Outlier-Aware Threshold Neuron to handle extreme activations via dual sub-neurons, and the Hierarchically Gated Neuron to approximate nonlinear functions through hierarchical decomposition approximation.
    \item  \textbf{Spike-equivalent LLM Component.} We present a fully spike-based LLMs by converting all key components into spike-equivalent modules, including self-attention, feed-forward networks, layer normalization, and softmax.
    \item  \textbf{SOTA Results on Eight LLMs.} We validate the proposed LAS method on both language and vision-language tasks. Notably, on the large OPT-66B model LAS surpasses the performance of vanilla model by 2\% in WSC task.
\end{itemize}

\section{Related Works}


\subsection{Spiking Neurons for ANN-to-SNN conversion}
The Integrate-and-Fire (IF) neuron~\cite{cao2015spiking} has dominated implementations of ANN-SNN conversion method due to its theoretically established equivalence with ReLU activations under rate coding schemes~\cite{rueckauer2017conversion,bu2023optimal}. This characteristic makes IF neurons particularly computationally efficient for implementing ReLU-based model. Additionally, the Leaky Integrate-and-Fire (LIF)~\cite{teeter2018generalized}
neurons improve robustness by adding a leakage mechanism to prevent infinite potential accumulation. Subsequent studies have successfully applied these two neurons to CNNs~\cite{Diehl2015FastclassifyingHS, rueckauer2016theory,Deng2021OptimalCO,Li2022EfficientAA,hao2024lm}. However, these neurons inherently based on linear dynamics that fundamentally limit their capacity to process nonlinear and non-monotonic functions, e.g., GELU. This intrinsic limitation severely restricts their compatibility with Transformer-based LLMs where such nonlinearities are ubiquitously employed. In contrast, the Few Spikes (FS) neuron \cite{stockl2021optimized} employs temporal coding with parameterized spike dynamics, which can effectively emulate non-monotonic activation functions over few time steps. Nevertheless, when apply FS to LLM, a primary issue is the presence of activation outliers, which
enlarge the quantization step sizes and subsequently will cause significant accuracy loss.

\subsection{ANN-to-SNN conversion for Transformer}
The Transformer architecture primarily relies on attention mechanisms and nonlinear operations like softmax, LayerNorm, and activation functions, which are challenging to convert directly into spiking forms. For example, SpikeZIP-TF ~\cite{you2024spikezip} aligns activation-quantized Transformer ANN with SNNs. ECMT ~\cite{huang2024towards} preserves nonlinear expectations via an Expectation Compensation Module and optimizes spike communication using multi-threshold neurons. Nevertheless, both fail to convert nonlinear operations into spike. 
SpikedAttention ~\cite{hwang2024spikedattention} introduces trace-driven matrix multiplication and a winner-oriented spike shift to implement spike-based softmax but struggles with LayerNorm and GELU activations. STA~\cite{jiang2024spatio} approximates nonlinear operations via Universal Group Operators and addresses non-causal interactions with Temporal-Corrective Self-Attention, yet requires $\ge 256$ time steps for conversion. All existing methods are limited to small vision transformers and overlook challenges in large generative language models. In contrast, LAS successfully converts all module and achieves ANN-comparable performance on OPT-66B with only 16 time steps.

\section{Preliminary}
FS neuron is a variation of the standard spiking neuron model. 
Unlike conventional spiking models, it employs fixed temporal parameters $\theta(t)$ (threshold), $h(t)$ (reset strength), and $d(t)$ (output weight) across $T$ time steps to approximate the activation function $f(x)$ of its ANN counterpart. This approximation is realized by aggregating weighted spikes $\hat f(x) = \textstyle \sum_{t=1}^T d(t)s(t)$, where $s(t) \in \{0,1\}$ denotes the binary spike state at timestep $t$.  

The dynamics of neuron begin with an initial membrane potential $v(1) = x$, where $x$ is the gate input. At each timestep $t$, the membrane potential updates according to  
\begin{equation}  
v(t+1) = v(t) - h(t)s(t),  
\end{equation}  
which exist a reset mechanism modulated by $h(t)$ after spike emission. A spike $s(t) = 1$ is fire when the membrane potential exceeds the threshold $\theta(t)$:  
\begin{equation}
    s(t)
= \Theta\bigl(v(t)-\theta(t)\bigr)
= \Theta\!\Bigl(x-\textstyle\sum_{j=1}^{t-1}h(j)\,s(j)-\theta(t)\Bigr),
\quad t=1,\dots,T.
\end{equation}
where $\Theta(\cdot)$ represents the Heaviside step function.  By optimizing the parameters $\{\theta(t), h(t), d(t)\}$, the FS neuron emulates the target activation $f(x)$ with few time step.


\section{Methodology}

\begin{figure}[ht] \centering  
\includegraphics[width=1\columnwidth]{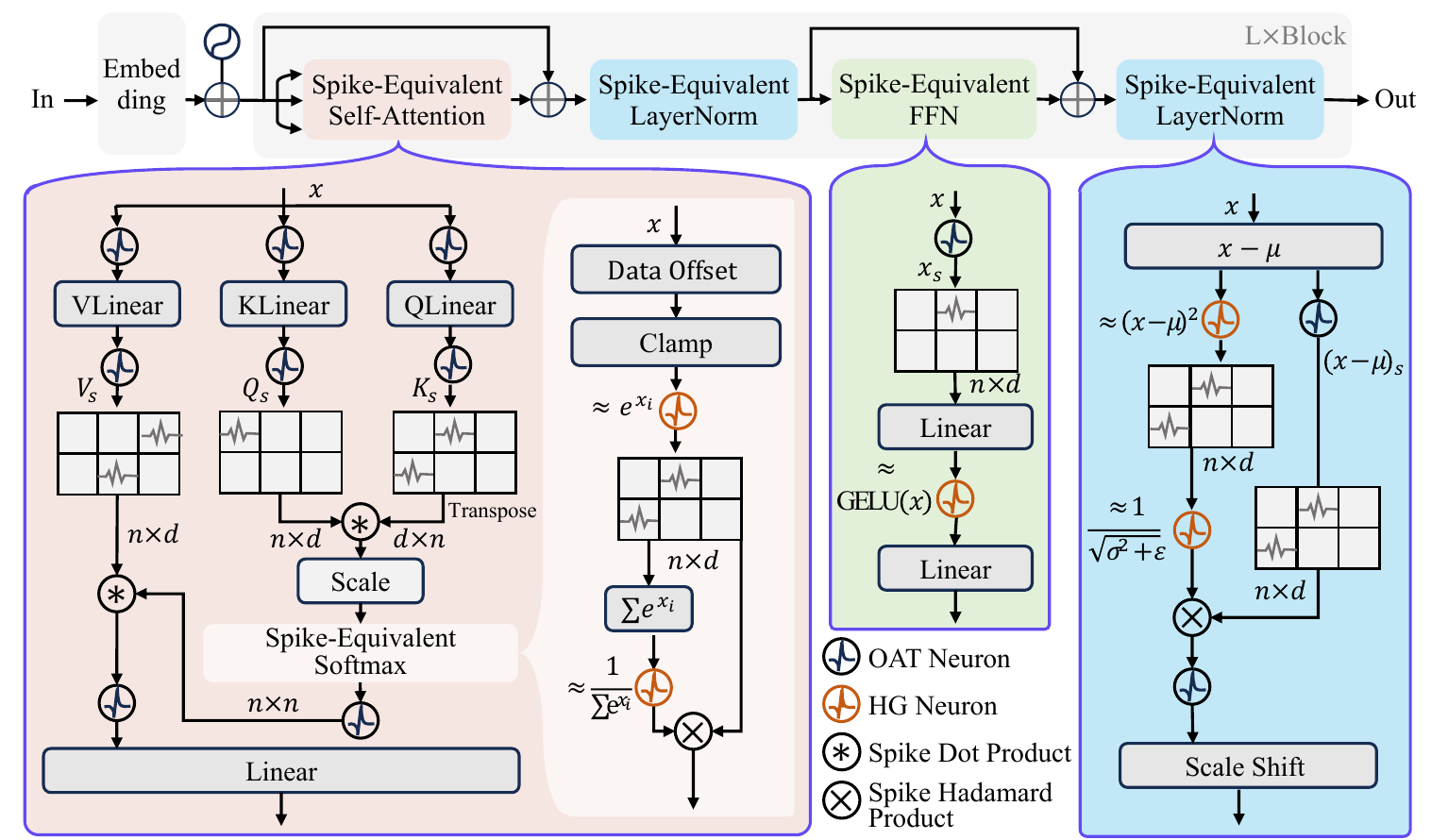} 
\caption{The overview of the proposed LAS method. OAT and HG neurons are designed to convert activation outliers and nonlinear operations of ANN-based LLMs, respectively. $n$ and $d$ denote the number of tokens and the channel dimensions, respectively.}

\label{mainFg}
\end{figure}

The framework of the proposed LAS method is illustrated in Figure~\ref{mainFg}. The Transformer-based LLM can be converted to fully spike-driven LLMs by using the proposed Outlier-Aware Threshold (OAT) and Hierarchically Gated (HG) neurons. More specifically, we insert the OAT neuron before every linear layer and matrix operation to deal with the outliers of LLMs. Moreover, the HG neuron is developed to simulate the nonlinear functions of LLM components.

\subsection{Spike Neurons Tailored for Spiking LLMs}
\paragraph{OAT neuron.}
To reduce the energy consumption of LLMs, we introduce a spiking neuron before every linear layer and matrix operation, thereby converting floating-point computations into low-power spike events. However, the outliers of LLMs enlarge the activation range, causing single spiking neuron compresses most values into the same bin, resulting severe information loss. Moreover, the bipolar nature of activations (positive and negative) challenges single-threshold schemes in capturing the full dynamic range.  To overcome this, we propose the OAT neuron, which comprises two Multi-Threshold (MT) sub-neurons that separately process normal and outlier activations. Each MT neuron employs multiple thresholds to handle positive and negative activation efficiently, reducing energy consumption and latency while maintaining representational fidelity.



Concretely, let $\mathbf{v}(1)\in\mathbb{R}^n$ be the vector of input membrane potentials at time step 1, which serves as gate input. the OAT neuron dynamics follow:
\begin{equation}
    \mathcal{M}_{\mathrm{out}} = \Theta\bigl(   \left |\mathbf{v}(1) \right |    -  \theta _{\mathrm{nor}}\bigr),\quad
\mathcal{M}_{\mathrm{nor}} = \mathbf{1} - \mathcal{M}_{\mathrm{out}},
\end{equation}
\begin{equation}
    s_i = \begin{cases} 
    \operatorname{MT-N}_{\text{nor}}(v_i{(1)}), &  \mathcal{M}_{\mathrm{out}} = 1 \\
    \operatorname{MT-N}_{\text{out}}(v_i{(1)}), &  \mathcal{M}_{\mathrm{nor}} = 1
    \end{cases}
\end{equation}
Here, $ \Theta\bigl(\cdot)$ is the Heaviside function. The normal threshold $\theta _{\text{nor}}$ determines the binary masks $\mathcal{M}_{\mathrm{out}}$ and $\mathcal{M}_{\mathrm{nor}}$. $\operatorname{MT-N}(\cdot)$ is function of MT neuron. $\operatorname{MT-N}_{\text{nor}}(\cdot)$  processes normal activations using $\theta_{nor}$, while $\operatorname{MT-N}_{\text{out}}(\cdot)$ handles outlier activations with a distinct threshold $\theta_{out}$ (where $\theta_{out}>\theta_{nor}>0$). Finally, $s_i$ is the output spike.

Each MT neuron builds on FS neuron, augmented by multiple thresholds to encode more information within a single time step. At time step $t$, we set $\theta(t) = h(t) = d(t) = \tau \cdot 2^{-t} $, where \(\tau\) is normal or outlier threshold. So that the neuron implements a coarse‐to‐fine approximation of a continuous  activation. We equip the neuron with symmetric positive/negative base thresholds $\pm\theta$ and $2H$ discrete threshold levels. At each time step, the membrane potential $v(t)$ selects the nearest available threshold for spike generation and potential reset,the membrane and spike dynamics  of MT neuron follow :
\begin{equation}
    v(t)=v(t-1)-h(t)z(t),
\end{equation}
\begin{equation}
    s(t) = \begin{cases}
        1,  & \left | v(t) \right |  \ge  \theta  \\
        0, & otherwise 
    \end{cases},
\end{equation}


\begin{equation}
d(t)=\begin{cases}
 \frac{2H-1}{H}\theta(t), &  v(t)\ge 2\theta(t)  \\
 \frac{H+k}{H}\theta(t), & \frac{H+k}{H} \theta(t)  \le v(t)< \frac{H+k+1}{H} \theta(t),k=0,1,...,H-1 \\
 0, & otherwise \\
 -\frac{H+k}{H}\theta(t), & -\frac{H+k}{n} \theta(t)  < v(t)\le  -\frac{H+k+1}{H} \theta(t),k=0,1,...,H-1\\
-\frac{2H-1}{H}\theta(t), & v(t)\le -2\theta(t)
\end{cases}
\end{equation}


The MT neuron achieves significant efficiency improvements over conventional spiking approaches. Where rate-coded neurons require $N$ time steps to represent $N$ distinct values, our binary coding scheme reduces this to $\frac{1}{n}\log_2(N)$ time steps.

\paragraph{HG neuron.}
FS neurons can approximate arbitrary nonlinear functions with sparse spikes. However, their approximation error for nonlinear functions is substantially amplified when processing activation outliers in LLMs.
To address this, we introduce the HG neuron, a neural unit composed of $N$ FS sub-neurons that together realize hierarchical approximation. For activation values within the hierarchy $(\lambda_{i-1}, \lambda_i]$, we allocate the $i$-th FS sub-neuron to process them. Specifically, the allocation is managed by a gating mechanism, defined as :
\[
M_{i,j} =
\begin{cases}
1, & \lambda_{i-1} \,\le\, v_j(1) < \lambda_i\\
0, & \text{otherwise}
\end{cases},
\]
where \(\mathcal{M}_{i,j}\) denotes the mask for the \(i\)-th FS neuron on the \(j\)-th input activation. Neurons remain silent when their corresponding mask values are zero.  so that each FS neuron \(FS_i\) is then responsible for approximating the nonlinear transform \(f(\cdot)\) over its sub‐range:
\begin{equation}
    s_{i,j} \;=\; FS_i\!\bigl(v_j\bigr)\,\cdot\,M_{i,j}.
\end{equation}
The overall output of HG neuron  recombines these partial approximations as:
\begin{equation}
    \hat{f}\bigl(v_j\bigr) \;=\; \sum_{i=1}^{N} s_{i,j} .
\end{equation}
The threshold parameters $\lambda_i$ are dynamically adjusted according to the statistical distribution of activation values in pre-trained LLMs, ensuring that both typical and outlier ranges are covered efficiently. To enable each sub-neuron $FS_i$ to approximate the target function $f$ without access to real training data, we define a uniform distribution $D$ over the interval $(\lambda_{i-1}, \lambda_i]$ and draw $M$ samples $\{x_j\}$ from $D$ so as to cover all possible inputs in that range. The resulting synthetic dataset $\{(x_j, f(x_j))\}$ serves as our training data. 


\subsection{Spike-Equivalent LLM Components}
\subsubsection{Spike-Equivalent Self-Attention}
Self-attention is the key component of Transformer architectures. We introduce Spike-Equivalent Self-Attention, which reformulates conventional self-attention using three spiking-friendly primitives: Spike Activation–Weight (SAW) Multiplication, Spike Activation–Activation (SAA) Multiplication, and Spike-Equivalent Softmax (detailed in Section~\ref{subsec:spike_softmax}).
\paragraph{SAW Multiplication.}
The input spike trains are projected via fixed weight matrices to produce spiking queries, keys, and values. Concretely, let $W\in\mathbb{R}^{n\times d}$ be a fixed weight matrix and variable features $X$, we can conclude that:
\begin{equation}
    Q = W \cdot X = W \cdot \sum_{t=1}^{T}\theta(t)X_s(t)  =\sum_{t=1}^{T} W \cdot \theta(t)X_s(t) 
\end{equation}
where  \(X_s(t)\in\{0,1\}^d\) is the binary spike input at time step \(t\), and \(\theta(t)\) is scalar threshold. The product \(W \cdot v(t)X_s(t)\) serves as weighted spike output for each time step. The final output is obtained by accumulating these responses over all time steps.

\paragraph{SAA Multiplication.} 
This operation is performed between dynamically generated spike-based matrices. Taking the dot-product attention between queries and keys as an example, the spike-based attention score can be expressed as:
\begin{equation}
\label{eqAA}
    A_T = Q_s \cdot K_s = \sum_{t=1}^{T}\theta_q(t)Q_s(t) \sum_{t=1}^{T}\theta_k(t)Q_k(t) =\sum_{i,j=1}^{T} \theta_q(i)\theta_k(i)Q_s(j)Q_k(j) 
\end{equation}
where $A_T$ denotes the attention score matrix accumulated over $T$ time steps, which is equivalent to ANNs.
To compute the expected matrix product output incrementally in SNNs, we decompose the calculation at each time step $t$ as follows:
\begin{equation}
    A_s(t) = \theta_q(t)\theta_k(t)Q_s(t)K_s(t) + \theta_q(t)Q_s(t)S_k(t)+S_q(t)\theta_k(t)K_s(t),
\end{equation}
where $S_q(t)=\sum_{i=1}^{t-1}\theta_q(i)\,K_s(i)$ and  $S_k(t)=\sum_{i=1}^{t-1}\theta_k(i)\,Q_s(i)$, which is the accumulated spikes of query and key. \(\theta_q(t)\theta_k(t)\) serves as the spike weight, and the computation only used the binary operations. This design can avoid costly multiplications and enables efficient, incremental spike-based attention computation over time. The detailed proof is provided in the \textbf{Appendix}~\ref{sec:appendix_SSA}.

\subsubsection{Spike-Equivalent Feed-Forward Network}
Conventional Feed-Forward Networks (FFNs) consist of two fully connected layers separated by a non-linear activation function. To reduce energy consumption, we replace all floating-point operations with discrete spike events. This is achieved by first converting the input to each fully connected layer into binary spike trains via the OAT neuron, and then approximating the activation function using the HG neuron. Formally, the spike-equivalent FFN is defined as:
\begin{equation}
    \mathrm{FFN}(x)
= \hat{f}\bigl(\phi (x)\,W_{1} + b_{1}\bigr)\,W_{2} + b_{2},
\end{equation}
where $\phi(\cdot)$ denotes the OAT neuron, and $\hat{f}(\cdot)$ denotes the HG neuron that approximates the activation function. Both components accept floating-point inputs and emit binary spike outputs, ensuring that the entire FFN operates  through spike events without any floating-point arithmetic.
An advanced variant, the gated FFN, which has demonstrated improved performance, is detailed in \textbf{Appendix}~\ref{sec:appendix_gateFFN}.

\subsection{Approximation for Non-Linearity}

To address the mismatch between the high-dimensional input of operations like LayerNorm and Softmax and the unary processing nature of HG neuron, we decompose these operations into the simpler, spike-compatible primitives, and apply HG neurons to approximate nonlinear functions. 


\paragraph{Spike-Equivalent LayerNorm.}
\label{subsec:spike_layernorm}
We propose a spike-compatible variant of LayerNorm by separating the standard mean–variance normalization and inverse square root scaling into two stages, both implemented with spike event. The transformation for an input \(x_i\) is defined as:
\begin{equation}
   \hat{\text{LN}}(x_i) = \gamma \cdot \! \phi(  \phi(X_{\mathrm{i}}-\mu ) \circ\ \hat{f}_{\mathrm{InvSqrt}}(\sigma^2)) + \beta\approx  \gamma \cdot \left( \frac{x_i-\mu}{\sqrt{\sigma^2+\epsilon}} \right) + \beta
\end{equation}
where $\circ$ represents the spike Hadamard product, following the same implementation as in Eq.\~(\ref{eqAA}), and \(\sigma^2\) is computed by squaring and summing \(x\), with the squaring operation itself approximated by HG neurons. The function \(\hat{f}_{\mathrm{InvSqrt}}(\cdot)\) employs HG neuron to approximate \(1/\sqrt{\sigma^2+\epsilon}\).  

\paragraph{Spike-Equivalent Softmax.}  
\label{subsec:spike_softmax}

The Softmax function for an input vector \(z\in\mathbb{R}^n\) is given by :
\begin{equation}
    \sigma_i(z) \;=\;\frac{\exp(z_i)}{\sum_{j=0}^{N-1}\exp(z_j)}
\;=\;\frac{\exp\!\bigl(z_i - z_{\max}\bigr)}{\sum_{j=0}^{N-1}\exp\!\bigl(z_j - z_{\max}\bigr)},
\end{equation} 
where \(z_{\max}\) is used to stabilize the exponential.  
This formulation consists of exponentiation, max-subtraction, and reciprocal normalization. Although the HG neuron can approximate the exponential and reciprocal functions, it cannot directly capture the dynamic subtraction of \(z_{\max}\). To address this, we reconstruct \(z_i - z_{\max}\) in the spike form. Let \(z_i(t)\) be the input of neuron \(i\) at time step \(t\).  We define a corrected spike output:  
\begin{equation}
    \hat{z}_i(t)
\;=\;z_i(t)
\;+\;\max_{0 \le m < N-1}\bigl(\sum_{j=1}^{t-1} z(j) \bigr)
\;-\;\max_{0 \le m \le N-1}\bigl(\sum_{j=1}^{t} z(j)  \bigr),
\end{equation}
where \(\hat{z}_i(t)\) is \(z_i - z_{\max}\) output at time step \(t\), the detailed derivation is provided in \textbf{Appendix}~\ref{sec:appendix_softmax}. Finally, we use HG neurons to approximate the remaining nonlinearities. Denoting  \(\hat{f}_{\exp}(\cdot)\) and \(\hat{f}_{\mathrm{inv}}(\cdot)\) as the HG neuron approximation of the exponential function and the reciprocal, respectively. The spike-equivalent output is computed as :
\begin{equation}
    \hat{\sigma}_i(z)
\;=\;\hat{f}_{\exp}\!\bigl(\hat{z}_i\bigr)
\;\circ\;\hat{f}_{\mathrm{inv}}\!\Bigl(\sum_{j=0}^{N-1}\hat{f}_{\exp}\!\bigl(\hat{z}_j\bigr)\Bigr).
\end{equation} 

This design implements Softmax normalization in a fully event-driven manner, making it compatible with neuromorphic accelerators.


\section{Experiments}
\subsection{Experimental Setup}
To evaluate our method, we converted pre-trained BERT-base, the OPT family (2.7B–66B), GPT-2, LLava1.5-7B, and Qwen2-VL-7B into spiking LLMs with 16 time steps. 
We then assessed language understanding on GLUE and  zero-shot reasoning on PIQA, ARC, OpenBookQA, Winogrande, COPA, WSC, and RTE. 
We assessed language generation on Enwik8 (bits per byte) and WikiText-103 (perplexity).
Finally, We measured multimodal performance on ScienceQA, RealWorldQA, BLINK, POPE, HallusionBench, MMStar, and MME. Additional details are provided in \textbf{Appendix}~\ref{sec:appendix_exp_detail}.

\begin{table*}[!ht] 
\centering
\small
\caption{
Comparing the accuracy of zero-shot tasks between LAS and SOTA OPT family.
}
\resizebox{1.0\textwidth}{!}{%
\begin{tabular}{llllllllll}
\toprule
\textbf{Model} & \textit{S} & \textit{T} & \textbf{PIQA} & \textbf{ARC} & \textbf{OpenbookQA} & \textbf{Winogrande} & \textbf{COPA} & \textbf{WSC} & \textbf{RTE} \\
\midrule
Flipped-11B~\cite{ye2022guess} & \ding{55} & N/A & 60.34& 30.81 & 17.60 & 57.85 & 67.00 & 65.57 & 52.71\\
T0-11B ~\cite{sanh2022multitask} & \ding{55} & N/A &73.67 & 68.39 & 29.00 & 62.98 & 81.00 & 75.09 & 84.48\\
Pythia-12b~\cite{biderman2023pythia}& \ding{55} & N/A & 76.00&70.08 & 26.60 & 63.54 &84.00 & 81.68 & 55.60\\
GPT-NeoX-20B~\cite{black2022gpt} & \ding{55} & N/A & 75.8&72.43 &  29.60 & 66.30 & 85.00 &83.52&  57.76\\
BLOOM-176B~\cite{le2023bloom}& \ding{55} & N/A & 	77.00 & 75.93 & 47.2& 67.00& 84.00& -& 57.4\\
\midrule
OPT-2.7B \cite{zhang2022opt} & \ding{55} & N/A & 73.78 & 60.73 & 25.00 & 61.33 & 77.00 & 78.02 & 55.25 \\
LAS (OPT-2.7B) & \ding{51} & 16 &73.61 &61.24 &   24.40  & 60.62  & 78.00 & 78.02 & 54.97\\
\midrule
OPT-7B \cite{zhang2022opt} & \ding{55} & N/A & 76.26 & 65.57 & 27.60 & 65.43 & 81.00 & 82.05 & 55.25 \\
FAS (OPT-7B) & \ding{51} & 16 & 73.23 & 64.73 & 27.00 & 60.38 & 83.00 & 77.66 & 55.60 \\
LAS (OPT-7B)~\cite{chen2025fas}  & \ding{51}  & 16 &76.22 & 65.95 & 27.40 & 65.75 & 80.00 & 81.69 & 55.96\\
\midrule
OPT-13B \cite{zhang2022opt} & \ding{55} & N/A & 75.95 & 67.13 & 27.20 & 65.27 & 81.00 & 82.78 & 58.12 \\
LAS(OPT-13B)& \ding{51}  & 16  & 76.28 &  67.38  & 26.20  & 65.51  & 80.00 & 82.05   & 57.40 \\
\midrule
OPT-30B \cite{zhang2022opt} & \ding{55} & N/A & 77.58 & 70.03 & 30.20 & 68.35 & 82.00 &  82.42 &57.76 \\
LAS (OPT-30B)& \ding{51} & 16 & 77.80   & 70.24  & 30.80 & 68.35 & 82.00 & 81.32  & 57.76 \\
\midrule
OPT-66B \cite{zhang2022opt} & \ding{55} & N/A & 78.73 & 71.72 & 30.40 & 69.98 & 85.00 & 82.78 & 60.55 \\
\textbf{LAS (OPT-66B)}& \ding{51}  & 16  &\textbf{78.67}  &  \textbf{71.25} & \textbf{32.00} & \textbf{68.27}  & \textbf{85.00} & \textbf{85.71}   & \textbf{59.93}\\
\bottomrule
\end{tabular}%
}
\label{tab:tableOPT}
\end{table*}

\subsection{Overall Results}
\begin{wraptable}[17]{r}{0.5\columnwidth} 
  \centering
    \centering 
    \captionof{table}{Comparing LAS with SOTA GPT models on the NLG
dataset. ‘En8’ stands for Enwik8, with BPB as the metric. ‘WT’
is WikiText-103 using perplexity. The lower the better for both
metrics. } 
    \resizebox{0.5\textwidth}{!}{ 
    \begin{tabular}{lllll}
    \toprule
    \textbf{Model} & \textit{S} & \textit{T} &\textbf{En8} &\textbf{WT} \\
    \midrule
    GPT-2 \cite{radford2019language} & \ding{55} &N/A& 0.96 & 16.53 \\
    Transformer-SSA \cite{hussain2023information} &\ding{55} &N/A& 1.02 & 16.91 \\
    \midrule
    AstroSNN \cite{Shen2023AstrocyteEnabledAI} & \ding{51} &  $-^{**}$ & 1.14 & 32.97 \\
    spikeGPT \cite{zhu2023spikegpt} & \ding{51}&1024 & 1.26 & 18.01 \\ \hline
    SPR (GPT-2) \cite{hao2023reducing} & \ding{51} &32 ($16^\dagger$) &1.01 & 19.24 \\ 
    QCFS (GPT-2) \cite{bu2023optimal} & \ding{51} &32 &1.02 & 19.36 \\ 
    COS (GPT-2) \cite{Hao2023BridgingTG}& \ding{51} &16 ($16^\dagger$)&1.01 & 19.15 \\  
    FAS (GPT-2) & \ding{51} &{16 }&{0.97} & {16.84} \\  \midrule
    \textbf{Our (GPT-2)} & \ding{51} &\textbf{16 }&\textbf{0.97} & \textbf{16.79} \\ 
    \bottomrule
    \end{tabular}
    }
\label{tableGPT}
\end{wraptable}
\textbf{Experiments on NLG Tasks.} LAS achieves state-of-the-art performance across the OPT family and GPT-2 models, as shown in Tables~\ref{tab:tableOPT} and \ref{tableGPT}. On zero-shot tasks, LAS preserves or even improves accuracy across all OPT scales (from 2.7B to 66B). For instance, it surpasses the original OPT-66B on OpenbookQA with scores of 32.00 compared to 30.40, and on WSC with 85.71 versus 82.78, all using just 16 time steps. Notably, even though BLOOM-176B was evaluated in a one-shot setting, our 66B model outperforms it on four tasks, highlighting our LAS’s superiority. Furthermore, LAS consistently reflects the expected trend of increased accuracy with larger model scales, indicating faithful preservation of capabilities across sizes.
In GPT model, LAS matches GPT-2 on Enwik8 with a score of 0.97 and shows only a slight degradation on WikiText-103, while substantially outperforming existing direct training and ANN-SNN conversion methods.

\textbf{Experiments on NLU Tasks.} FAS achieves near-lossless conversion for language understanding tasks. As presented in  Table~\ref{tableBERT}, with 16 time steps, LAS reaches 92.55\% accuracy on SST-2, which closely matches the original BERT’s 92.66\%, and even surpasses the ANN by 0.02\% on QQP. It also significantly outperforms existing SNN models; for example, SpikingBERT achieves only 88.19\% accuracy on SST-2 despite using 60 time steps.
Notably, our method narrows the accuracy gap to under 0.1\% across all NLU tasks, demonstrating the effectiveness of our lossless conversion strategy.

\begin{table*}[!ht]  
\centering
\caption{
Comparing LAS with SOTA models of BERT on the GLUE evaluation set. \textit{S} denotes whether an SNN or not. \textit{T} is the time steps. $^*$ denotes non-convergence. $^\dagger$ indicates additional time steps required to gather the necessary prior information.   The three blocks group models of non-SNN, direct trained and ANN-SNN converted.
}
\resizebox{1.0\textwidth}{!}{
\begin{tabular}{llllllllll}
\toprule
\textbf{Model} & \textit{S} &\textit{T }&\textbf{QQP} & \textbf{MNLI-m} & \textbf{SST-2} & \textbf{QNLI} & \textbf{RTE} & \textbf{MRPC} & \textbf{STS-B} \\
\midrule
BERT \cite{Devlin2019BERTPO} & \ding{55} &N/A &90.71 & 84.11 & 92.66 & 90.99 & 64.98 & 84.56/89.19 &88.70/88.48 \\
CBoW \cite{Wang2018GLUEAM} &\ding{55} &N/A& 75.00 & 57.10 & 79.50 &  62.50 & 71.90 &  75.00/83.70 &70.60/71.10\\
BiLSTM \cite{Wang2018GLUEAM} & \ding{55}&N/A & 85.30 & 66.70 & 87.50 &  77.00 & 58.50 &  77.90/85.10 &71.60/72.00\\
BiLSTM + Attn, CoVe \cite{Wang2018GLUEAM}&\ding{55}  &N/A& 83.50 & 67.90  & 89.20 &  72.50 & 58.10 & 72.80/82.40 & 59.40/58.00\\
GenSen \cite{subramanian2018learning} &\ding{55} &N/A& 82.60 & 71.40  & 87.20 & 62.50 & 78.40 & 80.40/86.20 &81.30/81.80\\
\midrule
SNN-TextCNN \citep{Lv2023SpikingCN} &\ding{51} & 50 & \(0.00^\star \) & 64.91 & 80.91 & 64.91 & 47.29 & -/80.62 & \(0.00^\star \)/- \\
spikeBERT \cite{lv2024spikebertlanguagespikformerlearned} &\ding{51} &4 & 68.17 & 71.42 & 85.39 & 66.37 & 57.47 & -/81.98 & -/\(18.73^\star \) \\
SpikeLM \cite{xing2024spikelm} &\ding{51}& 4 & -&77.10&87.00&85.30&69.00&-/85.70&84.90/-   \\
SpikingBERT \cite{bal2024spikingbert} & \ding{51}&60 & 86.82 &  78.10 &  88.19 &  85.20 &  66.06 &  79.17/85.15 &  82.20/81.90\\
\midrule
SPR (BERT)\cite{hao2023reducing} & \ding{51}&8 ($16^\dagger $) &87.48&77.56 & 90.48&87.75&64.98&78.68/85.76&86.71/86.50    \\ 
QCFS (BERT)\cite{bu2023optimal}  & \ding{51}&8 &88.42&79.57&89.91&86.80&56.68&78.92/85.37&86.18/85.82\\ 
COS (BERT)\cite{Hao2023BridgingTG}  & \ding{51}&8 ($8^\dagger$) &88.85&79.91&89.79&87.37&63.18&79.66/86.33&86.49/86.23\\ 
FAS (BERT) & \ding{51} &4& 90.38  & 82.77 & 91.17 & 90.13 & \textbf{66.06} & \textbf{86.02/90.22} & 87.46/87.26 \\ \midrule
\textbf{LAS (BERT)}  & \ding{51} & 16 & \textbf{90.73} & \textbf{84.19}& \textbf{92.66} & \textbf{90.92}& 65.34 & 84.80/89.42& \textbf{88.76/88.53}\\
\bottomrule
\end{tabular}
}
\label{tableBERT}
\end{table*}

\begin{table}[htbp] 
  \centering
  \caption{Compare the performance of LAS and SOTA multimodal LLMs on  vision-language tasks.}
  \label{tabblemutimodel}
  \resizebox{1.0\textwidth}{!}{%
  \begin{tabular}{llllllll}
    \toprule
    Model & ScienceQA & RealWorldQA & BLINK & POPE & HallusionBench & MMStar & MME \\
    \midrule
    LLava1.5-7b         & 67.22 & 52.41 &  41.22 & 82.81  & 81.36  & 33.40 & 1732.41 \\
    Qwen2-VL-7b         & 83.09 & 67.18 & 51.55   & 85.32  & 130.66  & 53.73 & 2240.55 \\
    \midrule
    LAS (LLava1.5-7b)   & 66.38  &49.15    &  41.01   & 80.79 & 67.93  & 33.86  &1613.04  \\
    \textbf{LAS (Qwen2-VL-7b)}   & \textbf{81.40} & \textbf{66.79}   & \textbf{52.18}  &\textbf{ 84.77} & \textbf{125.81}  & \textbf{53.86}  &\textbf{2222.59} \\
    \bottomrule
  \end{tabular}
}
\end{table}

\begin{table}[ht]
\centering
\caption{Accuracy and Energy Consumption of BERT under Different H Values}
\label{tableEnergy}
\resizebox{1.0\textwidth}{!}{
\begin{tabular}{c c c c c c c c c c}
\hline
\multirow{2}{*}{\textbf{Model}} & \multirow{2}{*}{\textbf{Metric}} & \multirow{2}{*}{\textbf{Original (ANN)}} & \multicolumn{7}{c}{\textbf{Our (SNN)}} \\
\cline{4-10}
 & & & H=1 &H=3 & H=5 & H=7 & H=10 & H=12 & H=15 \\
\hline
\multirow{2}{*}{BERT} & acc & 88.70  & 88.73  & 88.80  & 88.79  & 88.77 & 88.76  & 88.79  & 88.78 \\
\cline{2-10}
 & energy (\%) & 1 & 1.03 & 0.63 & 0.48 & 0.50 & 0.41 & 0.43 & 0.39 \\
\hline
\end{tabular}
}
\end{table}



\textbf{Experiments on Vision-Language Tasks.} As shown in Table~\ref{tabblemutimodel}, LAS demonstrates great performance with only minimal degradation. On Qwen2-VL-7B, LAS achieves scores of 66.79 on RealWorldQA and 84.77 on POPE, closely matching the original ANN model, and even outperforming it on BLINK and MMStar. Notably, although the LLaVA1.5 model has the same parameter size as Qwen2-VL-7B, it is inferior compared to Qwen2-VL-7B. The proposed LAS method still preserves this performance gap, indicating that the quality of the pre-trained ANN significantly impacts the performance of the resulting SNN. This is the potential limitations of LAS and underscores the importance of selecting high-quality pre-trained LLMs as the foundation for SNN conversion.

\begin{table*}[!ht]
\centering
\caption{
Comparing  BERT with different Time step.
}
\label{timestepEXP}
\resizebox{1.0\textwidth}{!}{
\begin{tabular}{llllllllll}
\toprule
\textbf{Model} & \textit{S} &\textit{T }&\textbf{QQP} & \textbf{MNLI-m} & \textbf{SST-2} & \textbf{QNLI} & \textbf{RTE} & \textbf{MRPC} & \textbf{STS-B} \\
\midrule
BERT \cite{Devlin2019BERTPO} & \ding{55} &N/A &90.71 & 84.11 & 92.66 & 90.99 & 64.98 & 84.56/89.19 &88.70/88.48 \\
\midrule
\multirow{4}{*}{\textbf{LAS(BERT)}} & \ding{51} & 10 & 68.64 &32.96 & 49.08 & 49.73& 47.29 & 68.63/81.29& 20.58/26.53\\
& \ding{51}& 11 & 86.29 &71.41 & 79.36 & 86.14& 47.65 & 76.96/82.06& 79.56/82.51\\
& \ding{51}& 13 & 90.69 & 84.11& 92.55 & 90.91& 66.06 & 84.31/89.12& 88.78/88.60\\
&\ding{51} & 16 & \textbf{90.73} & \textbf{84.19}&\textbf{ 92.55} & \textbf{90.92}& \textbf{65.34} & \textbf{84.80/89.42}& \textbf{88.79/88.58}\\
\bottomrule
\end{tabular}
}
\end{table*}

\subsection{Energy Analysis}

We first compare the energy consumption of nonlinear operations. A native GELU implementation requires approximately 70 FLOPs per activation due to the exponents in tanh. Jiang et al.~\cite{jiang2024spatio} introduce a Universal Group Operator (UGO) to approximate GELU, reducing its computational cost by 59\%. In contrast, our HG neuron encodes the GELU nonlinearity using at most 16 spikes, reducing the energy cost to near zero while maintaining high fidelity.


We then evaluate the overall energy efficiency of our SNN on the STS-B task by measuring its energy consumption relative to the original BERT model across discrete threshold levels \(H = 1\) to \(12\) in MT neuron (Table~\ref{tableEnergy}). At \(H = 1\), the energy ratio is 1.0319, showing a slight 3.19\% increase. Efficiency improves quickly with \(H\): the ratio drops to 0.7109 at \(H = 2\) (28.91\% reduction), 0.4818 at \(H = 5\) (51.82\%), and reaches 0.4147 at \(H = 10\) (58.53\%). Beyond \(H = 5\), the ratio remains below 0.50, such as 0.4417 at \(H = 12\), indicating stable and substantial energy savings at moderate to high discrete threshold levels. The detailed energy estimation methods are provided in \textbf{Appendix}~\ref{sec:appendix_energy_est}.


\subsection{Parameter and Efficacy Studies}
\begin{figure}[!t]
\begin{minipage}[t]{0.32\textwidth} 
\centering
\includegraphics[width=\linewidth]{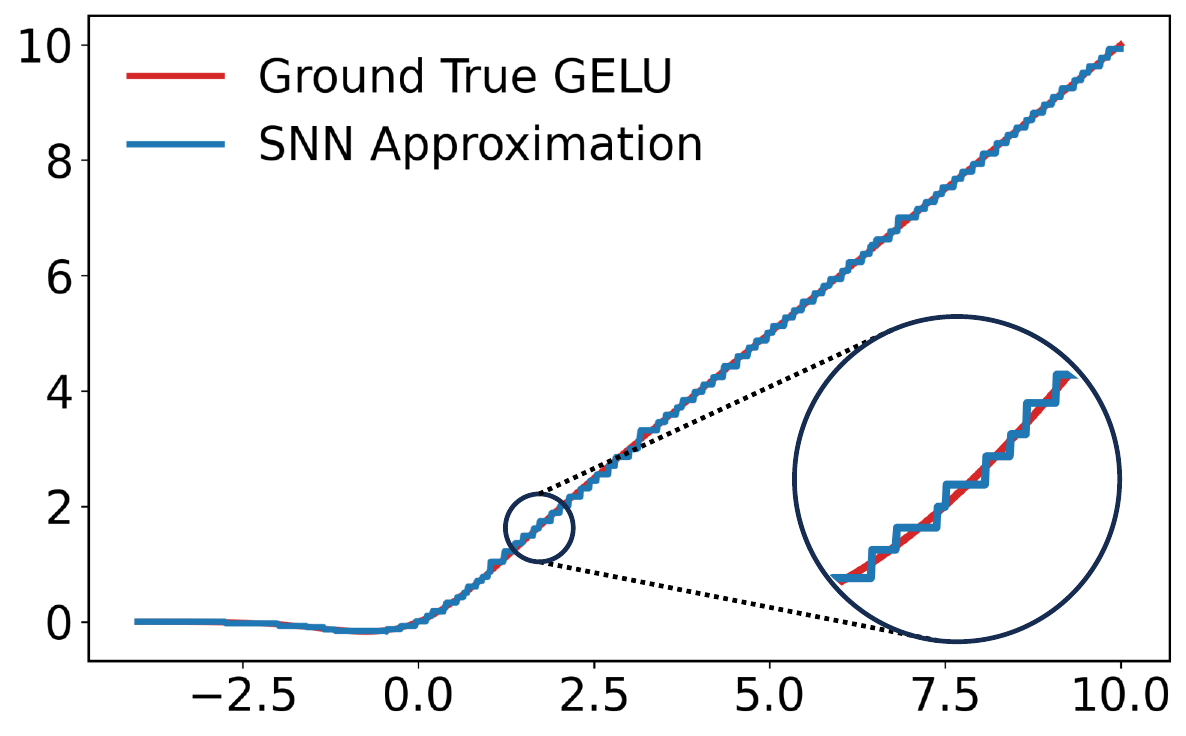}
  \caption{An approximated for GELU with time step=16.}
  \label{geluPic}
\end{minipage}
\hfill
\begin{minipage}[t]{0.32\textwidth} 
\centering
\includegraphics[width=\linewidth]{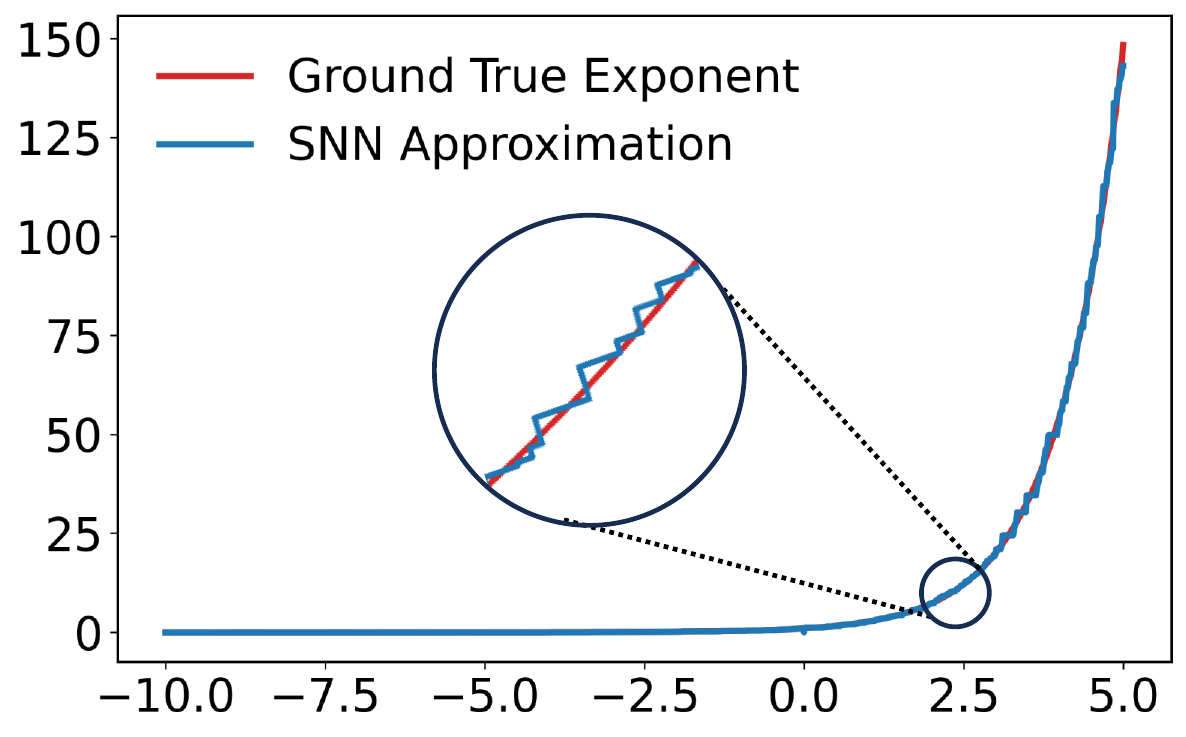}
  \caption{An approximated for exponent with time step=16.}
  \label{expPic}
\end{minipage}
\hfill
\begin{minipage}[t]{0.32\textwidth} 
\centering
\includegraphics[width=\linewidth]{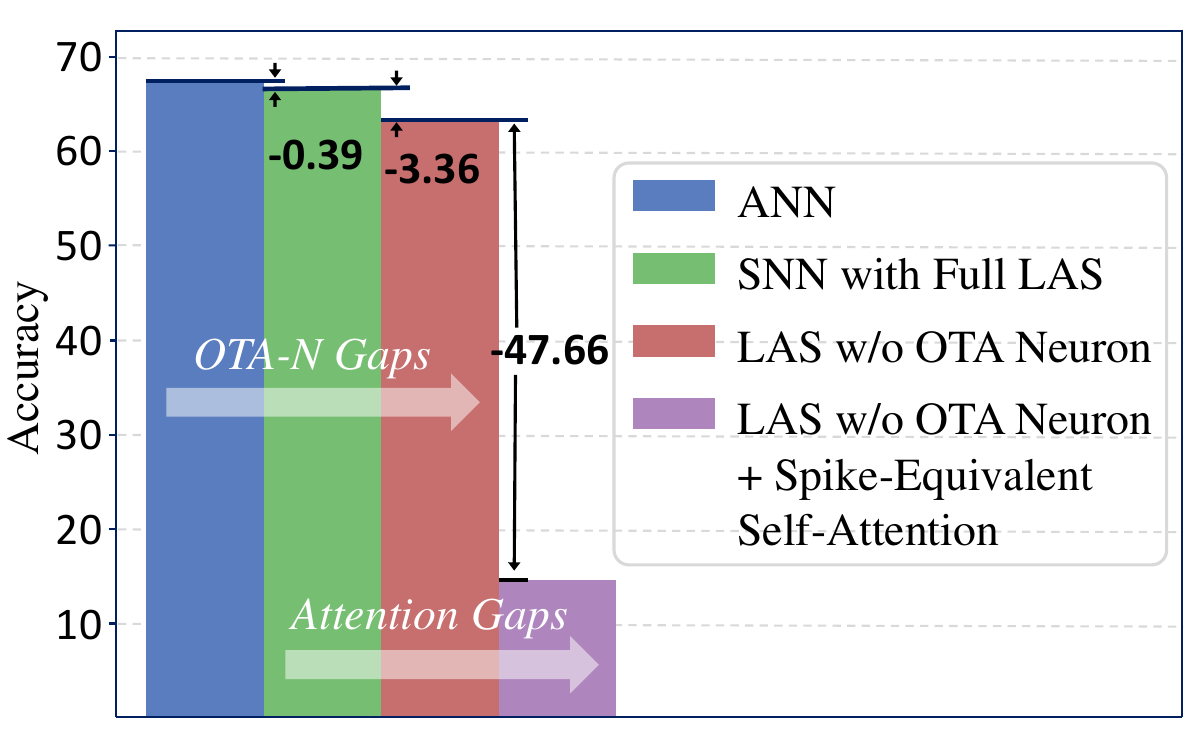}
\captionof{figure}{Ablations on components in STSB task} 
\label{fig:Ablation}
\end{minipage}
\end{figure}

\paragraph{Parameter Study on Time Steps.}
We evaluated the BERT model converted to an SNN across varying numbers of time steps, as shown in Table~\ref{timestepEXP}. The results reveal a strong, nonlinear dependence on the timestep count. At 16 time steps, the spiking BERT closely matches the original, achieving 92.55\% on SST-2 compared to 92.66\%, and 88.79/88.58 on STS-B compared to 88.70/88.48—demonstrating virtually lossless conversion. Even at 13 time steps, performance remains stable, maintaining 92.55\% on SST-2. However, at 11 time steps, performance degrades significantly, dropping to 79.36\% on SST-2 and 47.65\% on RTE. At the extreme case of 10 time steps, the model fails catastrophically, reaching only 49.08\% on SST-2 and a Spearman score of 26.53 on STS-B. These findings suggest that 13 to 16 time steps are sufficient for LAS to accommodate activation outliers and nonlinear dynamics, whereas fewer time steps result in irreversible information loss.
\paragraph{Efficacy study of HG neuron}  
To validate the ability of the Hierarchically HG neuron to approximate common nonlinear functions, we conducted experiments on the GELU and exponential functions. As shown in Fig.~\ref{geluPic}, HG neuron closely matches the overall shape and key transition points of the GELU curve; similarly, in Fig.~\ref{expPic}, it achieves an excellent fit to the $\exp(x)$ function. In both cases, the output of HG neuron and the original function curves almost perfectly overlap, demonstrating that HG neuron can achieve high-fidelity approximation of complex nonlinear operations.

\subsection{Ablation Study}

We conducted comprehensive ablation experiments on the RealWorldQA benchmark using the Qwen2-VL-7B model to quantify the impact of each major component in LAS. As shown in Figure~\ref{fig:Ablation}, our full LAS implementation achieves an accuracy of 66.79\%, nearly matching the original ANN with only a 0.39\% gap. Replacing the OAT neuron with a single MTN neuron (i.e., removing the OAT neuron) causes a 3.36\% drop in accuracy, highlighting its crucial role in preserving information fidelity by processing normal and outlier activations through dual sub-neurons. The most dramatic decline occurs when we disable our spike‑equivalent self‑attention: accuracy plunges to just 15.77\%, a 47.66 point decrease, demonstrating that this mechanism is essential for maintaining Transformer‑style contextual reasoning. Together, these results confirm that both the OAT neuron and the spike‑equivalent attention mechanism are critical for achieving high‑fidelity conversion of LLMs.

\section{Conclusion}
This paper proposes a Loss-less ANN-SNN conversion method for fully spike-driven large language models, termed LAS. Specifically, by introducing two specialized neurons that address activation outliers and nonlinear operations, LAS can transform all floating-point computations of ANN-based LLMs into energy-efficient spike computations. Moreover, the proposed spike-equivalent modules for self-attention, feedforward layers, Softmax function, and layer normalization further eliminate performance degradation. Experiments demonstrate SOTA performance of LAS across language understanding, generation, and multimodal reasoning tasks with only 16 time steps, achieving near-lossless conversion for models up to 66B parameters. To the best of our knowledge, it is the first time obtaining the high-performance and fully spike-driven LLMs with such a model size.




\bibliographystyle{plain}

\bibliography{neurips_2025}

\newpage
\appendix

\section{Derivation of SAA Multiplication}
\label{sec:appendix_SSA}

This operation is performed between dynamically generated spike-based matrices. Taking the dot-product attention between queries and keys as an example, the spike-based attention score can be expressed as:
\begin{equation}
    A_T = Q_s \cdot K_s = \sum_{t=1}^{T}\theta_q(t)Q_s(t) \sum_{t=1}^{T}\theta_k(t)Q_k(t) =\sum_{i,j=1}^{T} \theta_q(i)\theta_k(i)Q_s(j)Q_k(j) 
\end{equation}
where $A_T$ denotes the attention score matrix accumulated over $T$ time steps, which is equivalent to ANNs.
To compute the expected matrix product output incrementally in SNNs, we decompose the calculation at each time step $t$ as follows:
\begin{equation}
\begin{aligned}
A_S(t) &= \sum_{i=1}^{t}A(i) - \sum_{i=1}^{t-1}A(i)\\
&= A_t - A_{t-1} \\
&=\sum_{i=1}^{t} \theta _q(i) Q_s(i)\sum_{i=1}^{t} \theta _k(i) K_s(i) -\sum_{i=1}^{t-1} \theta _q(i) Q_s(i)\sum_{i=1}^{t-1} \theta _k(i) K_s(i)\\
&=\theta _q(t)\theta _k(t)Q_s(t)K_s(t) + \theta _q(t)Q_s(t) \cdot  \sum_{i=1}^{t-1} \theta  _k(i)Q_k(i)  + \sum_{i=1}^{t-1}\theta  _q(i)Q_s(i) \cdot  \theta _k(t)K_s(t) \\
&= \theta_q(t)\theta_k(t)Q_s(t)K_s(t) + \theta_q(t)Q_s(t)S_k(t) + S_q(t)\theta_k(t)K_s(t) 
\end{aligned}
\end{equation}
where $S_q(t)=\sum_{i=1}^{t-1}\theta_q(i)\,Q_s(i)$ and  $S_k(t)=\sum_{i=1}^{t-1}\theta_k(i)\,K_s(i)$.

\section{Spiking Gated Feed-Forward Network}
\label{sec:appendix_gateFFN}
\begin{figure}[ht] \centering  
\includegraphics[width=0.8\columnwidth]{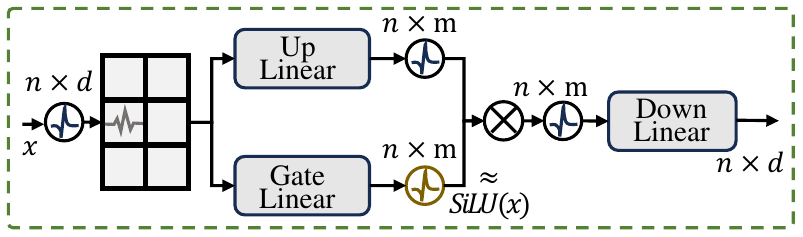} 

\caption{The overview of the spiking gate FFN. }

\label{fig:GateFFN}
\end{figure}

The gated FFN, or gated MLP, is a variant of the conventional FFN used in Transformer architectures. Unlike standard FFNs that apply a single activation function between two linear projections, gated MLPs introduce a multiplicative interaction between two projected vectors, one of which is modulated by a nonlinear activation (the gate). 

Formally, the gated MLP transforms an input vector $x \in \mathbb{R}^{n\times d}$ through the following steps:
\begin{align}
    g &= \sigma(W_g x + b_g) && \text{(Gate projection with activation)} \nonumber \\
    u &= W_u x + b_u && \text{(Up projection)}  \nonumber \\
    z &= u \odot g && \text{(Element-wise gating)} \\
    f(x) &= W_d z + b_d && \text{(Down projection)} \nonumber
\end{align}
where $W_{\text{gate}}, W_{\text{up}} \in \mathbb{R}^{d \times m}$ project the input $x \in \mathbb{R}^{n \times d}$ to an intermediate space, and $W_{\text{down}} \in \mathbb{R}^{m \times d}$ recovers the original dimension. The bias vectors $b_g$ and $b_u$ are in $\mathbb{R}^{m}$, and $b_d$ is in $\mathbb{R}^{n}$. The activation function $\sigma(\cdot)$ is typically chosen to be  SiLU, and the symbol $\odot$ denotes Hadamard product.

In this work, we extend the gated MLP into a spike-equivalent form by introducing an OAT neuron before each linear layer to convert floating-point inputs into spike. Additionally, the nonlinear activation is replaced with spike events by HG neurons. As illustrated in Figure \ref{fig:GateFFN}, this design completely eliminates floating-point computation entirely and enables fully event-driven processing, while preserving the expressive power of multiplicative gating.  Formally, the spike-equivalent gated MLP can be described as:
\begin{align}
    g &= \hat{f}(W_g \phi(x) + b_g) && \text{(Gate projection with activation)} \nonumber \\
    u &= W_u \phi(x) + b_u && \text{(Up projection)}  \nonumber \\
    z &= \phi(u) \circ g && \text{(Element-wise gating)} \\
    f(x) &= W_d\phi(z) + b_d && \text{(Down projection)} \nonumber
\end{align}
where $\phi(\cdot)$ denotes the OAT neuron, and $\hat{f}(\cdot)$ denotes the HG neuron that approximates the activation function SiLU. \(\circ\) represents the spike Hadamard product.

\section{Derivation of Spike Offset in Softmax} 
\label{sec:appendix_softmax}
The Softmax function for an input vector $z \in \mathbb{R}^n$ is defined as: :
\begin{equation}
    \sigma_i(z) \;=\;\frac{\exp(z_i)}{\sum_{j=0}^{N-1}\exp(z_j)}
\;=\;\frac{\exp\!\bigl(z_i - z_{\max}\bigr)}{\sum_{j=0}^{N-1}\exp\!\bigl(z_j - z_{\max}\bigr)},
\end{equation} 
where subtracting $z_{\max}$ from each element stabilizes the exponential terms.

To reconstruct the stabilized term $z_i(t) - z_{\max}(t)$ in a spike-based equivalent form, we define $\hat{Z}_{i,T}$ as the accumulated output of the $i$th activation over $T$ time steps in the SNN ,which approximates the output of the corresponding input activation in the ANN :
\begin{equation}
    \hat{Z}_{i,T}=\sum _{t=1}^Tz_{i} - max(\sum_{t=1}^TZ).
\end{equation}

So, we can calculation the output at time step $t$:

\begin{equation}
\begin{aligned}
\hat{z}_i(t) 
&= \sum_{i=1}^{t} \hat{z}_i(i) - \sum_{i=1}^{t-1} \hat{z}_i(i) \\
&= \hat{Z}_{i,t} - \hat{Z}_{i,t-1} \\
&= \sum_{k=1}^t z_i(k) - \max\left(\sum_{k=1}^t z(k)\right) - \left( \sum_{k=1}^{t-1} z_i(k) - \max\left(\sum_{k=1}^{t-1} z(k)\right) \right) \\
&= z_i(t) + \max\left(\sum_{k=1}^{t-1} z(k)\right) - \max\left(\sum_{k=1}^{t} z(k)\right) \\
\end{aligned}
\end{equation}
where \(\hat{z}_i(t)\) is \(z_i - z_{\max}\) output of SNN at time step \(t\).

\section{Experimental Details}
\label{sec:appendix_exp_detail}
\subsection{Datasets}
 For  Natural Language Understanding (NLU) tasks, we chose seven different types of tasks, i.e., six classification and one regression tasks, from the GLUE benchmark. We selected Quora Question Pair (QQP) and Microsoft Research Paraphrase Corpus (MRPC) for classification tasks, and Semantic Textual Similarity Benchmark (STSB) for regression task to evaluate our LAS on similarity and paraphrase tasks. For inference tasks, we opted for MultiGenre Natural Language Inference (MNLI), Question Answering NLI (QNLI), and Recognizing Textual Entailment (RTE) datasets. For single-sentence-based sentiment analysis tasks, we chose Stanford Sentiment Treebank (SST-2). Accuracy is the metric for QQP, MNLI-m, SST-2, QNLI, RTE. MRPC combines accuracy and F1 scores. STS-B uses the Pearson/Spearman correlation.

For NLG task, we chose the following two classic text classification datasets, i.e., Enwik8 and WikiText-103, to evaluate the text generation performance of LAS. Specifically, the Enwik8 dataset is a large-scale text dataset consisting of the first 100 million characters from Wikipedia. It is widely used for character-level language modeling and text generation tasks, providing a challenging benchmark for models due to its extensive and varied content. The Bit-Per-Byte (BPB) metric is commonly employed to assess its performance.
In addition, the WikiText-103 dataset is another comprehensive text dataset derived from Wikipedia articles. It contains over 100 million words and is known for its high-quality, naturally occurring text. WikiText-103 is commonly used for training and evaluating language models, particularly in tasks involving text generation, language modeling, and machine translation. Perplexity (PPL) is the metric of choice for evaluating the performance.

We evaluate vision–language models on seven benchmarks spanning multimodal reasoning, spatial understanding and domain‐specific perception. ScienceQA integrates images, diagrams and textual explanations to tackle science questions, while RealWorldQA emphasizes geometric spatial reasoning in real‐world scenarios such as autonomous driving. BLINK challenges holistic perception by combining object detection, OCR and commonsense reasoning, and POPE diagnoses object hallucination through controlled image–text alignment experiments. HallusionBench uses 346 curated images paired with 1 129 human‐crafted questions to expose language hallucination and visual illusion, MMStar presents 1 500 human‐selected challenge samples across six capability dimensions with novel Multi‐modal Gain and Leakage metrics, and the Multimodal Evaluation Benchmark assesses perception and cognition over 14 leak‐proof subtasks for fair model comparison.

\subsection{Experimental Setup}
We converted pre-trained LLMs including BERT-base, GPT-2, the OPT family (2.7 B to 66 B), LLaVA 1.5-7 B, and Qwen2-VL-7 B into spiking LLMs with 16 time steps. For the Outlier-Aware Threshold neuron we applied a multi-threshold scheme with $H=5$ discrete levels in all models. In the Hierarchically Gated neuron, the number of FS neurons $N$ was optimally tuned per neuron according to each model’s error tolerance. 

All experiments were carried out on eight RTX 3090 GPUs. We evaluated OPT models using the open-source lm-eval toolkit and vision-language tasks with VLMEvalKit. Additionally, since the ViT component accounts for only 0.3 B of the 7 B parameters in LLaVA 1.5-7 B yet has a significant impact on accuracy, we retained its analog weights.


\section{Energy Estimation}
\label{sec:appendix_energy_est}
In ANNs, the energy consumption of floating-point operations (\(FLOPs\)) with multiplication and accumulation (MAC) , remains constant within a defined network structure. In contrast, SNNs rely on synaptic operations (SOPs) with sparse accumulation (AC), where energy consumption varies depending on spike sparsity. To quantitatively evaluate energy savings, we adopt the energy estimation equation from \cite{rathi2020diet}:
\begin{equation}
    \frac{E_{snn}}{E_{ann}}=\frac{SOPs\cdot E_{AC}}{FLOPs\cdot E_{MAC}},
\end{equation}
where \(E_{MAC}= 4.6\)  and \(E_{AC}=0.9\). For common unary nonlinearities, such as GELU, the computation is significantly more expensive. Specifically, Computing the exponential term requires approximately \(FLOPs \approx20\)  ~\cite{nilsson2014hardware}, while  the square‐root term about \(FLOPs \approx12\)~\cite{liu2012power}. 
Due to exponents in tanh, a native unary non-linear operator like GELU implementation incurs \(FLOPs \approx70\) per activation~\cite{jiang2024spatio}.

\end{document}